\documentclass[11pt]{article}
\pdfoutput=1
\usepackage{amsmath,amsfonts,hyperref,graphicx,algpseudocode,algorithm,bm,arydshln,epsfig,subfigure,appendix,afterpage,mathtools,cite,overpic}
\usepackage{amssymb}
\usepackage{tikz,setspace}
\usepackage{titlesec}

\providecommand{\keywords}[1]{\textbf{\textit{Index terms---}} #1}
\titlespacing\section{0pt}{10pt plus 4pt minus 2pt}{2pt plus 2pt minus 2pt}
\titlespacing\subsection{0pt}{10pt plus 4pt minus 2pt}{2pt plus 2pt minus 2pt}
\titlespacing\subsubsection{0pt}{10pt plus 4pt minus 2pt}{2pt plus 2pt minus 2pt}

\begin{document}
\title{Inferring object rankings based on noisy pairwise comparisons from multiple annotators}
\author{Rahul~Gupta,
        Shrikanth~Narayanan}% <-this % stops a space
%\IEEEcompsocitemizethanks{\IEEEcompsocthanksitem M. Shell is with the Department
%of Electrical and Computer Engineering, Georgia Institute of Technology, Atlanta,
%GA, 30332.\protect\\
%E-mail: see http://www.michaelshell.org/contact.html
%\IEEEcompsocthanksitem J. Doe and J. Doe are with Anonymous University.}% <-this % stops a space
%\thanks{}
\begin{spacing}{0.95}

\begin{abstract}
Ranking a set of objects involves establishing an order allowing for comparisons between any pair of objects in the set.
Oftentimes, due to the unavailability of a ground truth of ranked orders, researchers resort to obtaining judgments from multiple annotators followed by inferring the ground truth based on the collective knowledge of the crowd.
However, the aggregation is often ad-hoc and involves imposing stringent assumptions in inferring the ground truth (e.g. majority vote).
In this work, we propose Expectation-Maximization (EM) based algorithms that rely on the judgments from multiple annotators and the object attributes for inferring the latent ground truth. 
The algorithm learns the relation between the latent ground truth and object attributes as well as annotator specific ``probabilities of flipping'', a metric to assess annotator quality.
We further extend the EM algorithm to allow for a variable ``probability of flipping" based on the pair of objects at hand.
We test our algorithms on two data sets with synthetic annotations and investigate the impact of annotator quality and quantity on the inferred ground truth.
We also obtain the results on two other data sets with annotations from machine/human annotators and interpret the output trends based on the data characteristics.
\end{abstract}
\keywords{Learning to Rank, Expectation Maximization, Multiple Annotators, Support Vector Ranker}

\maketitle

\section{Introduction}
\label{sec:intro}
Given a set of items, ranking involves establishing a partial order over the items.
This ordering allows comparison between two items, in which the first is either ranked higher, lower or equal to the second \cite{hullermeier2008label}.
This is commonly termed as a pairwise approach and has been investigated in relation to information retrieval \cite{liu2009learning}, ranking web pages \cite{haveliwala2002topic} and even analysis of human behavioral constructs such as emotions \cite{lin2008ranking}.
Within the problem of modeling preferences using pairwise comparisons, inferring the true order given comparisons from noisy annotators \cite{wu2011learning} is very relevant.
Often, due to the unavailability of the ground truth, experimenters resort to accumulating judgments from multiple annotators and performing a fusion of their collective knowledge.
This trend has existed beyond learning to rank and has also been observed in classification and regression tasks \cite{yan2010modeling}.
Particularly within the domain of classification, several researchers have proposed novel ways of jointly modeling the annotators in inferring the latent ground truth \cite{raykar2010learning,audhkhasi2013globally}.
Although prior research has addressed similar problems within ranking, the methods enforce a specific structure (e.g., Borda count method, Nanson method \cite{van2000variants,niou1987note}) on annotator judgments in inferring the latent ground truth.
In this work, we present Expectation-Maximization (EM) \cite{moon1996expectation} based algorithms inspired from work in classification problems to infer the latent ground truth in ranking objects.
Through these algorithms, we not only aim to relax the ad-hoc constraints imposed in ground truth computation of preferences but also open up possibilities to integrate the existing approaches within ranking and classification addressing similar problems.

Given noisy pairwise preferences from multiple annotators, the proposed algorithms target to infer a single ground truth ranking while also computing a reliability metric for each of the annotators.
We assume the ground truth to be a latent variable that can be inferred not only based on the noisy pairwise comparisons from multiple annotators, but also the distribution of a set of attributes/features corresponding to the pair of items being compared.
We approach this problem using the Expectation-Maximization (EM) framework \cite{bishop2006pattern} and develop a Joint Annotator Modeling (JAM) scheme, inspired from existing literature in modeling multiple annotators \cite{raykar2010learning,audhkhasi2013globally}.
The JAM schemes assume that, given the set of attributes/features for a pair of objects, there exists a latent true preference order.
Furthermore, the annotators either retain or flip this preference order based on an annotator-specific reliability metric, the ``probability of flipping".
The JAM scheme initially learns the relationship between the attributes of the object pair and the latent ground truth as well as each annotator's ``probability of flipping".
The final inference on the preference ground truth is made jointly taking into account the model's belief based on the object attributes and the annotators' preferences.
We further modify the JAM scheme to allow for non-constant ``probability of flipping" based on the pair of objects at hand, termed as Variable Reliability Joint Annotator Modeling (VRJAM) scheme.
We compare the JAM and VRJAM schemes to existing methods such as majority voting and fusion after Independent Annotator Modeling (IAM) (similar to Borda count method \cite{van2000variants}).
We evaluate our models on two data sets with synthetic annotations to investigate the impact of annotator quality and quantity on our models.
We also evaluate our models on two other data sets with annotations from machines (ground truths available) and humans (ground truths not available). 
We interpret the outcomes of the models based on the data characteristics and suggest a few future directions.
In the next section, we provide a background of the relevant work, followed by the description of various methodologies for inferring latent true preference order from noisy annotator preferences.

\section{Previous work}
Several researchers have addressed the problem of learning to rank from pairwise comparisons with applications to a variety of domains.
In particular, works by H{\"u}llermeier and F{\"u}rnkranz et al. \cite{hullermeier2008label, furnkranz2003pairwise} provide a comprehensive background on preference learning using the pairwise approach.
%Cao et al. \cite{cao2007learning} present a transition from pairwise approach to a listwise approach in learning to rank.
%Several other researchers have attempted to  incorporate other machine learning topics within the framework of learning to rank.
Considering consolidation of other machine learning topics within the framework of ranking, Brinker et al. \cite{brinker2004active} and Long et al. \cite{long2010active} integrated active learning in ranking problems, Chu et al. \cite{chu2005extensions} provided an extension of Gaussian processes for ranking and He et al. \cite{he2004manifold} used manifold based ranking for image retrieval. 
Other notable works proposing novel methods and applications for ranking include learning to rank using non-smooth cost functions \cite{quoc2007learning}, the Mcrank algorithm \cite{li2007mcrank} and learning to rank with partially labeled data \cite{duh2008learning}.
Whereas several existing works have addressed other interesting flavors of learning to rank \cite{cao2007learning}, rank aggregation \cite{dwork2001rank} is possibly one of the most well studied fields under this domain.
A prominent setting under rank aggregation is learning a probability distribution centered around a single or a mixture of global rankings.
Several works \cite{ding2015learning,gleich2011rank,pan2013rank} present algorithms for rank aggregation using non-negative matrix factorization, nuclear norm minimization and sparse decomposition techniques.

A different problem setting under learning to rank is inferring a ground truth ranking from a set of pairwise preferences available from multiple annotators. 
Chen et al. \cite{chen2013pairwise} address this problem and present an active learning framework that selects a pair of objects as well as the annotator to be queried while training a ranking model. 
Along similar lines, Kumar et al. \cite{kumar2011learning} investigated algorithms to fuse ranking models trained using noisy crowd.
The formulation of inferring latent ground truth from noisy annotations is particularly well studied in classification and regression problems. 
Dawid et al. \cite{dawid1979maximum} presented one of the earlier works in fusion annotator beliefs followed by more recent models by Raykar et al. \cite{raykar2010learning} and Zhou et al. \cite{zhou2012learning}.
Audhkhasi et al. \cite{audhkhasi2013globally} further extended the model to account for diversity in the reliability of annotators over the feature space.
Our algorithm carries similar goals as Chen el al. \cite{chen2013pairwise} and Kumar et al. \cite{kumar2011learning} to fuse preferences from multiple annotators, with modeling schemes inspired from proposals by Raykar et al. \cite{raykar2010learning} and Audhkhasi et al. \cite{audhkhasi2013globally}.
In the next section, we discuss the algorithms designed for the fusion of noisy pairwise comparisons from multiple annotators along with a few other baseline methods. 

\section{Methodology}

Given a set of $N$ items $\bm O = \{O_1,O_2,...,O_N\}$ and $K$ annotators, we represent the $k^\text{th}$ annotator's preference of $O_i$ over $O_j$ as $O_i^k \succ O_j^k$.
Our goal is to infer the latent ground truth denoted by $O_i \succ O_j$, indicating that $O_i$ is ranked higher than $O_j$.
We also assume the availability of attributes/feature values $\bm X = \{\bm x_1,\bm x_2,...,\bm x_N\}$ for each of the $N$ objects, where $\bm x_i$ is a vector of attributes for the item $O_i$.
We define the variables $z_{ij}^k$ $(k=1..K)$ and $z_{ij}^*$ to represent the preferences of the annotators and the ground truth as follows.

\vspace{-2mm}
\begin{equation} \label{membership_eq}
z_{ij}^* = \begin{cases}
  1  \text{ if}\ O_i \succ O_j \\
  0  \text{ if}\ O_j \succ O_i
\end{cases}
\hspace{-3mm}\text{and,}\;z_{ij}^k = \begin{cases}
  1  \text{ if}\ O_i^k \succ O_j^k \\
  0  \text{ if}\ O_j^k \succ O_i^k
\end{cases}
\hspace{-3mm},k=1..K
\end{equation}

Below we describe four methods to obtain the ground truth given the noisy pairwise comparisons between items. 
The first two methods, majority vote and Independent Annotator Model serve as a baseline.
Although fusion from these methods is easy to perform, they assume that each annotator is equally reliable in inferring the ground truth which may not always be the case. 
The next two methods, the Joint Annotator Modeling and Variable Reliability Joint Annotator Modeling schemes learns a reliability metric for each annotator.
The final decision is made based on available annotations as well as the attributes for the pair of objects at hand. 

\subsection{Majority Vote (MV)}
Majority voting is one of the most popular methods for merging decisions from multiple annotators and has been  consistently used in various classification experiments \cite{gupta2012classification,mower2011framework} as well as ranking \cite{kumar2011learning}.
In this method, we say that the inferred preference is $O_i \succ O_j$ if a majority of the annotators say so. In case of a tie among annotators, a random decision is taken between $z_{ij}^*=1$ and $z_{ij}^*=0$.
Note that this model does not use the object attributes $\bm X$ in inferring $z_{ij}^*$ and relies solely on $z_{ij}^k$ as shown in the graphical model in Figure \ref{all_models}(a).
Also, each annotator is weighted equally in deciding the majority.

\subsection{Independent Annotator Modeling (IAM)}
\label{sec:IAM}

In this scheme, we initially train annotator specific ranking models to capture the relation between object attributes and each annotator's preference rankings.
The ranking model for the $k^\text{th}$ annotator returns a score $f_k(\bm x_i)$ for every object $O_i$ based on the attributes $\bm x_i$.
Finally, the inferred ground truth value for $z_{ij}^*$ is given by comparing the sum of scores $f_k(\bm x_i)$ and $f_k(\bm x_j)$ over all the annotators ($k = 1..K$). 
This method is synonymous to the Bradley-Terry model \cite{rao1967ties} (extended by Chen el al. \cite{chen2013pairwise}) and the Borda count method \cite{van2000variants} used for aggregating decisions from multiple annotators.
In case of Bradley-Terry model, preference between two objects is determined based on their relevance scores, computed as a sum of $f_k(\bm x_j)$ over all the annotators in the current IAM scheme.
Similarly, in the Borda count method, each annotator scores every object and $z_{ij}^*$ is inferred by comparing sum of scores across all the annotators.
In this section, our substitute for the Borda count score for $O_i$, as given by the annotator $k$, is the value $f_k(\bm x_i)$.
We describe the model training and ground truth inference in detail below.\\

\noindent{\bf Training annotator specific models}:
Given the $k^\text{th}$ annotator's pairwise preferences $z_{ij}^k$, we train an annotator specific Support Vector Ranker (SVR) \cite{donmez2008optimizing} as the function $f_k$.
Our goal is to learn $\mathit f_k$ for every annotator $k$, such that the following holds.
\begin{equation}\label{pref_eq}
O_i^k \succ O_j^k \iff z_{ij}^k = 1 \iff \mathit f_k(\bm x_i) > \mathit f_k(\bm x_j)
\end{equation}

In this work, we chose $\mathit f_k$ to be a linear function characterized by a weight vector $\bm w_k$ such that $\mathit{f_k}(\bm x_i) =  \langle \bm w_k,  \bm x_i\rangle$, where $\langle \bm w_k, \bm x_i\rangle$ represents the dot product between $\bm w_k$ and $\bm x_i$.
An SVR targeting the problem in (\ref{pref_eq}) performs the following optimization on the cost function $\mathcal M_k$ \cite{donmez2008optimizing}.

\vspace{-3mm}
\begin{equation} \label{opt_cost}
\begin{aligned}
\bm w_k = \arg\min_{\bm w_k} \mathcal{M}_k &= \arg\min_{\bm w_k} \sum_{\substack{\text{All pairs } \\ \bm x_i, \bm x_j }} \hspace{-4mm}z_{ij}^k[1-\langle \bm w_k, \{\bm x_i - \bm x_j\} \rangle]_+ \\
+ &(1-z_{ij}^k)[1-\langle \bm w_k, \{\bm x_j - \bm x_i\} \rangle]_+
\end{aligned}
\end{equation}

In the equation above, $\{\bm x_i - \bm x_j\}$ depicts a notion of difference operator between $\bm x_i$ and $\bm x_j$ and $[\; ]_+$ represents the standard hinge loss function \cite{gentile1998linear}. 
In this work, we use $\{\bm x_i - \bm x_j\}$ to be a simple element-wise subtraction between attribute vectors $\bm x_i$ and $\bm x_j$.
We learn $\bm w_k$ ($\forall k = 1..K $) using the standard gradient descent algorithm \cite{burges2005learning}.
Since $\mathcal{M}_k$ is non differentiable, we use the approximation suggested by Rennie et al. \cite{rennie2005loss} in the hinge loss function. \\

\begin{figure}[tb]
\centering
\includegraphics[trim=0cm 6cm 0cm 0cm,clip=true,scale=.28]{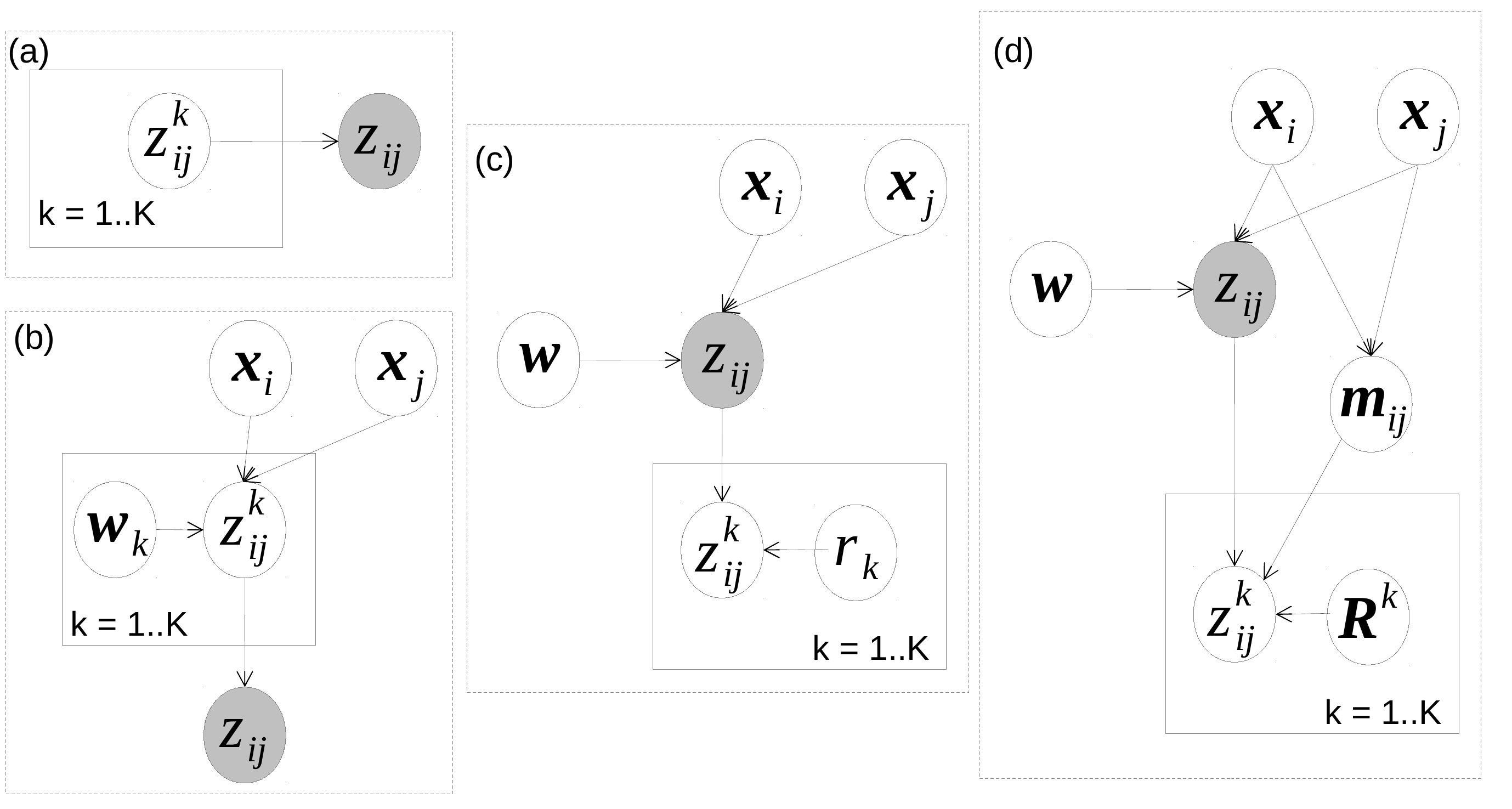}
\vspace{-2mm}
\caption{Graphical models for (a) Majority vote (MV) (b) Independent Annotator Model (IAM) (c) Joint Annotator Model (JAM) and, (d) Variable Reliability Joint Annotator Model (VRJAM) schemes. }
\vspace{-3mm}
\label{all_models}
\end{figure}

\noindent{\bf Fusing annotator models}:
After obtaining $f_k$ for each of the annotators, we say $z_{ij} = 1$ if: 
\begin{equation}
\sum_{k=1}^K f_k(\bm x_i) > \sum_{k=1}^K f_k(\bm x_j) 
\end{equation}
A graphical model representing this scheme is shown in Figure \ref{all_models}(b). 
Note in order to obtain $z_{ij}^*$, the scheme of unweighted combination is enforced on $f_k$ outputs.

\subsection{Joint Annotator Modeling (JAM)}
\label{sec:JAM}

In this section, we propose an Expectation-Maximization (EM) algorithm \cite{moon1996expectation} to infer the ground truth by jointly modeling the noisy comparisons.
Our algorithm is inspired by similar works \cite{raykar2010learning,audhkhasi2013globally} in the domain of classification problems.
A graphical model for this scheme is shown in Figure \ref{all_models}(c).
We assume the ground truth $z_{ij}^*$ to be a latent variable that can be inferred using the object attributes $\bm x_i$ and $\bm x_j$.
Our choice for inferring $z_{ij}^*$ based on $\bm x_i,\bm x_j$ is again an SVR model with a weight vector $\bm w$.
Furthermore, we assume that $z_{ij}^k$ is obtained by flipping the binary variable $z_{ij}^*$ with a probability $r_k$.
In summary, this model assumes that there is an inherent true preference given attributes from two objects and the annotators are flipping it based on annotator specific probabilities ($r_k,k=1..K$).
Consequently, the probability $r_k$ also provides a measure of annotator quality as a higher $r_k$ implies higher chances of an annotator committing an error.
We infer the latent ground truth $z_{ij}^*$ using an EM algorithm described in the next section.

\subsubsection{Expectation-Maximization algorithm}

The EM algorithm maximizes the log-likelihood $\mathcal L$ of the observed data, that is, annotator preferences given the object attributes and the model parameters.
In our case, $\mathcal L$ is given as shown in (\ref{em_ll1}).
Notice the introduction of the latent ground truth $z_{ij}^*$ into $\mathcal L$ in (\ref{em_ll2}). 

\begin{align} 
&\mathcal L = \sum_{\substack{\text{All pairs } \\ \bm x_i, \bm x_j}} \log p(z_{ij}^1, ..,z_{ij}^K/\bm x_i, \bm x_j, \bm w, r_1, ..,r_k) \label{em_ll1} \\
&=\hspace{-2mm} \sum_{\substack{\text{All pairs }\\ \bm x_i, \bm x_j}} \sum_{z_{ij}^*} \log p(z_{ij}^*,z_{ij}^1, ..,z_{ij}^K/\bm x_i, \bm x_j, \bm w, r_1, ..,r_k)  \label{em_ll2} 
\end{align} 

Following the EM derivation procedure in section 9.4 in \cite{bishop2006pattern}, we introduce a distribution over the latent ground truth $z_{ij}^*$: $q(z_{ij}^*)$. 
Consequently $\mathcal L$ can be written as sum of two terms, a Kullback Leibler (KL) divergence term $\text{KL}(q||p)$ and another log-likelihood term $\mathcal M$ as shown in (\ref{ll_split}).

\vspace{-2mm}
\begin{equation} \label{ll_split}
\mathcal L = \mathcal M + \text{KL}(q||p)
\end{equation}

where,
\begin{equation}
\begin{aligned}
&\mathcal M = \sum_{\substack{\text{all pairs } {\bm x_i,\bm x_j}}} \sum_{z_{ij}^*} q(z_{ij}^*) \times\\
& \log \Big\{\frac {p(z_{ij}^*,z_{ij}^1, ..,z_{ij}^K | \bm x_i, \bm x_j, \bm w, r_1, .., r_K)} {q(z_{ij}^*)} \Big\} 
\end{aligned}
\end{equation}

\begin{equation}
\begin{aligned}
&\text{KL} (q||p) = -\sum_{\substack{\text{all pairs } {\bm x_i,\bm x_j}}} \sum_{z_{ij}^*} q(z_{ij}^*) \times \\ 
&\log \Big\{\frac {p(z_{ij}^*|z_{ij}^1, ..,z_{ij}^K , \bm x_i, \bm x_j, \bm w, r_1, .., r_K)} {q(z_{ij}^*)} \Big\} 
\end{aligned}
\end{equation}

The EM algorithm consists of two steps: the E and M steps. 
In the E-step, $\mathcal M$ is maximized with respect to $q(z_{ij}^*)$ while holding the other parameters constant.
The solution is equivalent to the posterior distribution $p(z_{ij}^*|z_{ij}^1, ..,z_{ij}^K , \bm x_i, \bm x_j, \bm w, r_1, .., r_K)$.
In the M-step, $\mathcal M$ is maximized with respect to model parameters while holding the estimated distribution $q(z_{ij}^*)$ constant.
We describe the parameter initialization followed by the E and M steps below.

%After the initialization of model parameters ($\bm w,r_1,..,r_K$), the EM algorithm learns them by iteratively performing E and M-steps.
%In this model (please refer to Figure \ref{all_models}(c)), the E-step involves estimating the value of $z_{ij}$ and the M-step involves obtaining $\bm w$ and the probabilities $r_k, k=1..K$.
%We describe the EM steps below.\\

\noindent {\bf Initialization}: We randomly initialize the SVR weight vector $\bm w$ and the probabilities of flipping $r_k (k=1..K)$.\\

\noindent {\bf While} $\bm w,r_1,..,r_K$ not converged perform E and M-steps, where: \\

\noindent \underline{E-step}:
In the E-step, we set the probability distribution $q(z_{ij}^*)$ equal to $p(z_{ij}^*|z_{ij}^1, ..,z_{ij}^K , \bm x_i, \bm x_j, \bm w, r_1, .., r_K)$. 
This quantity can be represented as shown in (\ref{e_step_all}). 
A detailed derivation for this quantity can be seen in Appendix 1.

\vspace{-2mm}
\begin{equation}\label{e_step_all}
\begin{aligned}
&q(z_{ij}^*)=p(z_{ij}^*|z_{ij}^1, ..,z_{ij}^K , \bm x_i, \bm x_j, \bm w, r_1, .., r_K) =\\ 
&\Big(p(z_{ij}^*|\bm x_i, \bm x_j, \bm w) \times \prod_{k=1}^K p(z_{ij}^k | z_{ij}^*, r_k)\Big) / p(z_{ij}^1, ..,z_{ij}^K) 
\end{aligned}
\end{equation}

Note that the first term $p(z_{ij}^*|\bm x_i, \bm x_j, \bm w)$ in (\ref{e_step_all}) is conditioned on the SVR model parameters $\bm w$ and object attributes $\bm x_i$ and $\bm x_j$ only. 
Since SVR is not a probabilistic model, we apply a commonly used trick in support vector machine classifiers employed to obtain class probabilities.
The trick involves fitting logistic models to distance from the decision hyperplane to obtain the probabilities of preference decisions \cite{hastie1998classification} (A comparison of the hinge loss function and the logistic loss function is made in Appendix 3).
Equations (\ref{logistic_fit}) and (\ref{logistic_fit2}) show the computation for $p(z_{ij}^* |\bm x_i, \bm x_j, \bm w)$ using the logistic model. 
 
\begin{align}
&p(z_{ij}^* = 1|\bm x_i, \bm x_j, \bm w) = \frac{\exp{\langle \bm w, \{\bm x_i - \bm x_j\} \rangle}}{1 + \exp{\langle \bm w, \{\bm x_i - \bm x_j\}\rangle}} \label{logistic_fit}\\
&p(z_{ij}^* = 0|\bm x_i, \bm x_j, \bm w) = 1 - p(z_{ij}^* = 1|\bm x_i, \bm x_j, \bm w) \label{logistic_fit2} 
\end{align}

The second term $p(z_{ij}^k | z_{ij}^*,r_k)$ in (\ref{e_step_all}) is $r_k$ if $z_{ij}^k$ and $z_{ij}^*$ are in disagreement and $1-r_k$ otherwise, as shown below.

\vspace{-2mm}
\begin{equation}\label{flip_terms}
p(z_{ij}^k | z_{ij}^*,r_k) =
\begin{cases}
r_k \text{ if } z_{ij}^k \neq z_{ij}^*\\
1-r_k \text{ if } z_{ij}^k = z_{ij}^*
\end{cases} 
\end{equation}

Replacing the values in (\ref{e_step_all}) from (\ref{logistic_fit}) and (\ref{flip_terms}), we can represent $q(z_{ij}^* = 1)$ as shown in (\ref{estep}).
$q(z_{ij}^* = 0)$ can be computed accordingly.

\vspace{-2mm}
\begin{equation} \label{estep}
\begin{aligned}
&q(z_{ij}^* = 1) = \frac{\exp{\langle \bm w, \{\bm x_i - \bm x_j\} \rangle}}{1 + \exp{\langle \bm w, \{\bm x_i - \bm x_j\}\rangle}} \times\\ 
&\prod_{k=1}^K \hspace{-3mm} \underbrace{[(r_k)^{(1-z_{ij}^k)} \times (1-r_k)^{z_{ij}^k}]}_{\substack{r_k (/1-r_k) \text{is multiplied when $z_{ij}^k=0(/1)$} }} / p(z_{ij}^1, ..,z_{ij}^K)
\end{aligned}
\end{equation}

Note that the denominator $p(z_{ij}^1, ..,z_{ij}^K)$ is common between $q(z_{ij}^* = 1)$ and $q(z_{ij}^* = 0)$ and need not be computed.
We can just compute the numerator in (\ref{e_step_all}) for $q(z_{ij}^* = 1)$ and $q(z_{ij}^* = 0)$ and normalize these probabilities to sum to one.
In the next section, we discuss the M-step. \\

\noindent \underline{M-step}: 
In this step, we estimate the model parameters $\bm w, r_k (k=1..K)$ based on estimated distribution $q(z_{ij}^*)$.
These parameters are estimated by maximizing $\mathcal M$ after substituting $q(z_{ij}^*)$ estimated in the E-step.
In our case, $\mathcal M$ can be written as shown in (\ref{updated_M}). 
$\mathbb H(q(z_{ij}^*))$ is the entropy of $q(z_{ij}^*)$ and is a constant term with respect to the model parameters $\bm w, r_1, ..,r_k$.
We disregard the entropy term for further M-step derivations. 

\begin{equation} \label{updated_M}
\begin{aligned} 
&\mathcal M = \sum_{z_{ij}^*} q(z_{ij}^*) \times  
\log p(z_{ij}^*,z_{ij}^1, ..,z_{ij}^K| \bm x_i, \bm x_j, \bm w, r_1, .., r_K) \\ &+ \mathbb H(q(z_{ij}^*)) 
\end{aligned}
\end{equation}

We can rewrite $\mathcal M$ as shown in (\ref{updated_M2}). 
For a detailed derivation, please see Appendix 2.
Note that each parameter $\bm w$, $r_1,..,r_K$ appears in a separate term within the summation in (\ref{updated_M2}) and thus, we only need to consider the corresponding term while optimizing for a parameter.
We discuss the optimization for the SVR parameters $\bm w$ and flipping probabilities $\bm r_k$ below. 

\begin{equation} \label{updated_M2}
\begin{aligned} 
\mathcal M = \sum_{z_{ij}^*} q(z_{ij}^*) 
\Big(\sum_{k=1}^K \log p(z_{ij}^k| z_{ij}^*, r_k) + \log p(z_{ij}^*| \bm x_i, \bm x_j, \bm w)\Big)  
\end{aligned}
\end{equation}

\noindent{\it Obtaining SVR weight vector $\bm w$}: 
We only need to consider the following term $\mathcal M_{\bm w}$ within $\mathcal M$ to optimize for $\bm w$.

\begin{equation}
\mathcal M_{\bm w} =  \sum_{z_{ij}^*} q(z_{ij}^*)\log p(z_{ij}^*| \bm x_i, \bm x_j, \bm w) 
\end{equation}

In the EM algorithm, $\log p(z_{ij}^*| \bm x_i, \bm x_j, \bm w)$, would be obtained from a probabilistic model to infer $z_{ij}^*$ conditioned on $\bm x_i, \bm x_j,\bm w$.
However, since the choice of our model is a non-probabilistic SVR, we instead solve the following optimization in (\ref{em_approx_p}) to obtain $\bm w$.
We would like to point out that this is an approximation we use in the EM algorithm.
Appendix 3 shows the probability distribution corresponding to the logistic models used in (\ref{logistic_fit}) and its relation to the following optimization.

\begin{equation}\label{em_approx_p}
\begin{aligned}
\bm w = \arg\min_{\bm w} \mathcal M^\prime_{\bm w} &= \arg\min_{\bm w} \Big(q(z_{ij}^* = 1) [1-\langle \bm w, \{\bm x_i - \bm x_j\} \rangle]_+  \\
&+ q(z_{ij}^* = 0) [1-\langle \bm w, \{\bm x_j - \bm x_i\} \rangle]_+ \Big) 
\end{aligned}
\end{equation}

Note that $\mathcal M^\prime_{\bm w}$ in (\ref{em_approx_p}) is similar to the cost function $\mathcal L_k$ defined in (\ref{opt_cost}) for training annotator specific models.
However, instead of being trained on binary decisions values (e.g., $z_{ij}^k$ used in $\mathcal{L}_k$), $\mathcal M_{\bm w}$ is defined over the soft estimate $q(z_{ij}^*)$.
Next, we discuss the optimization problem to obtain $r_1,..,r_K$.

\noindent{\it Obtaining probability of flipping $r_1,..,r_K$}:
In order to obtain $r_k$, we need to optimize the following term within $\mathcal M$.

\begin{equation}
\begin{aligned}
&r_k = \arg\min_{r_k} \mathcal M_{r}^k \\ 
&= \arg\min_{r_k} \sum_{\text{All pairs } \bm x_i, \bm x_j} \sum_{z_{ij}^*} q(z_{ij}^*) \log p(z_{ij}^k | z_{ij}^*,r_k)
\end{aligned}
\end{equation}

$p(z_{ij}^k | z_{ij}^*,r_k)$ is replaced in the above equation as shown in (\ref{flip_terms}) and the term can be optimized to obtain $r_k$. We obtain the final inference for $z_{ij}^*$ as discussed below. \\

\noindent {\bf Final inference}: After convergence, we make the final inference on $z_{ij}^*$ based on obtained distribution $p(z_{ij}^*|z_{ij}^1, ..,z_{ij}^K , \bm x_i, \bm x_j, \bm w, r_1, .., r_K)$, as was derived in (\ref{e_step_all})-(\ref{estep}).
$z_{ij}^*$ is inferred to be 1 or 0 as per the following equation.  
\begin{equation}
\begin{aligned}
p(z_{ij}^* = 1|z_{ij}^1, ..,z_{ij}^K , \bm x_i, \bm x_j, \bm w, r_1, .., r_K) \substack{\overset{1}{>} \\ \underset{0}{<}} \\ 
p(z_{ij}^* = 0|z_{ij}^1, ..,z_{ij}^K , \bm x_i, \bm x_j, \bm w, r_1, .., r_K)
\end{aligned}
\end{equation}
 
%In the next section, we propose a modification to the present algorithms where the probability of flipping is considered variable and determined based on the difference vector $\{\bm x_j - \bm x_i\}$.
Next, we propose a modification to this scheme considering the annotators' probability of flipping to be variable.

\subsection{Variable Reliability Joint Annotator Modeling (VRJAM)}

This scheme is similar to the joint annotator model proposed in the previous section except for the probability of flipping $r_k$ being variable.
The motivation behind this scheme is that annotators may have variable reliability depending upon the pair of objects $O_i$ and $O_j$ at hand (a similar assumption is in the model proposed by Audhkhasi et al. \cite{audhkhasi2013globally}).
Therefore, instead of a constant $r_k$ for the annotator $k$, we determine a vector $\bm R_k = [r_k^1,..,r_k^D]$, where based on the difference vector $\{\bm x_i - \bm x_j\}$, one of the values $r_k^d (d=1..D)$ is chosen as the probability of flipping.
We retain the assumption that $z_{ij}^*$ is a latent variable conditioned on $\bm x_i, \bm x_j$ and the SVR weight vector $\bm w$.
We again train this model using an EM algorithm described below. The algorithm is similar to the EM algorithm in section \ref{sec:JAM} and we borrow several steps for the sake of brevity.

\subsubsection{Expectation-Maximization algorithm}

For the purpose of our experiments, we divide the space spanned by difference vectors $\{\bm x_i - \bm x_j\}$ into $D$ clusters.
For the $k^\text{th}$ annotator, a distinct probability of flipping $r_k^d (d=1..D)$ is computed in each cluster.
We obtain the clusters by performing the standard K-Means clustering \cite{hartigan1979algorithm} on the values $\{\bm x_i - \bm x_j\}$ obtained over all pairs $\bm x_i, \bm x_j \in \bm X$.
The membership of $\{\bm x_i - \bm x_j\}$ to a cluster is denoted by a $1$-in-$D$ encoding vector $\bm m_{ij} = [m_{ij}^1,..,m_{ij}^D]$ where $m_{ij}^d=1$ indicates that $\{\bm x_i - \bm x_j\}$ belongs to the $d^{\text{th}}$ cluster.
The overall graphical model for this scheme is represented in Figure \ref{all_models}(d).
The graphical model is very similar to the one in Figure \ref{all_models}(c), except for $\bm m_{ij}$ now determining the flipping probability.
The data log-likelihood term $\mathcal L$ in (\ref{em_ll1}) changes slightly to incorporate $\bm R_1, ..,\bm R_K$ and $\bm m_{ij}$ (instead of scalar values $r_1, ..,r_K$) as represented by $\mathcal L^\prime$ in (\ref{em_ll1_vrjam}). 
We perform the initialization, the E and M-steps and final inference as discussed in the next section. 

\vspace{-2mm}
\begin{equation} 
\hspace{-2mm}\mathcal L^\prime = \sum_{\substack{\text{All pairs } \\ \bm x_i, \bm x_j}} \log p(z_{ij}^1, ..,z_{ij}^K/\bm x_i, \bm x_j, \bm w, \bm R_1, .., \bm R_k, \bm m_{ij}) \label{em_ll1_vrjam} 
\end{equation} 

\noindent {\bf Initialization}: We randomly initialize the SVR weight vector $\bm w$ and the vectors $\bm R_k$ for all the annotators.
We perform K-means clustering to segment the space spanned by $\{\bm x_i - \bm x_j\}, \forall \bm x_i,\bm x_j \in \bm X$. The number of clusters $D$ is set empirically by gradually increasing $D$ until the distance between two cluster centroids falls below a threshold (compared to distances to other centroids). \\

\noindent {\bf While} $\bm w,\bm R_1,..\bm R_K$ not converged perform E and M-steps, where: \\

\noindent \underline{E-step}: 
The E-step is same as the E-step in section \ref{sec:JAM}.
The only difference is that $p(z_{ij}^k| z_{ij}^*, r_k)$ in (\ref{e_step_all}) is replaced by $p(z_{ij}^k| z_{ij}^*, \bm R_k, \bm m_{ij})$, which equals to the quantity in (\ref{flip_terms2}).
$\langle\bm R_k,\bm m_{ij}\rangle$ represents a dot product between $\bm R_k$ and $\bm m_{ij}$ to select an entry in $\bm R_k$ based on the cluster index corresponding to $\{\bm x_j - \bm x_i\}$.

\vspace{-2mm}
\begin{equation}\label{flip_terms2}
p(z_{ij}^k | z_{ij}^*,\bm R_k, \bm m_{ij}) =
\begin{cases}
\langle \bm R_k, \bm m_{ij} \rangle \text{, if } z_{ij}^k \neq z_{ij}^*\\
1-\langle \bm R_k, \bm m_{ij} \rangle \text{, if } z_{ij}^k = z_{ij}^*
\end{cases} 
\end{equation}

Consequently, $q(z_{ij}^*=1)$ is computed as shown in (\ref{estep_vrjam}). 
After estimating $q(z_{ij}^*)$, we estimate the model parameters as discussed next.
 
\vspace{-3mm}
\begin{equation} \label{estep_vrjam}
\begin{aligned}
&q(z_{ij}^* = 1) = \frac{\exp{\langle \bm w, \{\bm x_i - \bm x_j\} \rangle}}{1 + \exp{\langle \bm w, \{\bm x_i - \bm x_j\}\rangle}} \times\\ 
&\prod_{k=1}^K \hspace{-3mm} \underbrace{[\langle \bm R_k, \bm m_{ij} \rangle^{(1-z_{ij}^k)} \times (1-\langle \bm R_k, \bm m_{ij} \rangle)^{z_{ij}^k}]}_{\substack{\langle \bm R_k, \bm m_{ij} \rangle (/1-\langle \bm R_k, \bm m_{ij} \rangle) \text{is multiplied when $z_{ij}^k=0(/1)$} }} / p(z_{ij}^1, ..,z_{ij}^K)
\end{aligned}
\end{equation}

\noindent \underline{M-step}: In the M-step, we re-estimate the parameters $\bm w$ and the vectors $\bm R_k$. 
Value of $\mathcal M$ also alters in this formulation to incorporate $\bm R_1,.., \bm R_K$ and $\bm m_{ij}$.
$p(z_{ij}^k | z_{ij}^*,r_k)$ in (\ref{updated_M2}) is replaced by $p(z_{ij}^k | z_{ij}^*,\bm R_k, \bm m_{ij})$.
This has no impact on the estimation of $\bm w$, which remains the same as in section \ref{sec:JAM}. 
We describe the estimation of the vector $\bm R_k$ below.\\

\noindent{\it Obtaining probability of flipping entries in $\bm R_k$}:
The optimization framework to obtain $\bm R_k$ is shown below.
\begin{equation}
\bm R_k = \arg\min_{\bm R_k}\sum_{\substack{\text{All pairs } \\ \bm x_i, \bm x_j}} \sum_{z_{ij}^*} q(z_{ij}^*) \log p(z_{ij}^k | z_{ij}^*,\bm R_k, \bm m_{ij})
\end{equation}

The above optimization over the vector $\bm R_k$ can easily be broken down into scalar optimization over each of its entries after replacing $p(z_{ij}^k | z_{ij}^*,\bm R_k, \bm m_{ij})$ as shown in (\ref{flip_terms2}). 
We next discuss the final step for inferring $z_{ij}^*$.\\ 

\noindent {\bf Final inference}: The final inference on $z_{ij}^*$ is made based the following likelihood comparison once the model converges. 
This inference is similar to one in the JAM scheme. 

\begin{equation}
\begin{aligned}
p(z_{ij}^* = 1|z_{ij}^1, ..,z_{ij}^K , \bm x_i, \bm x_j, \bm w, \bm R_1, .., \bm R_K, m_{ij}) \substack{\overset{1}{>} \\ \underset{0}{<}} \\ 
p(z_{ij}^* = 0|z_{ij}^1, ..,z_{ij}^K , \bm x_i, \bm x_j, \bm w, \bm R_1, .., \bm R_K, m_{ij})
\end{aligned}
\end{equation}

In the next section, we evaluate various fusion schemes on several datasets with synthetic annotations as well as annotations obtained from machines and humans.

\section{Experimental Results}
We test the discussed ranking algorithms on two synthetically created data sets and two real world data set as discussed next. 
%In the next section, we present our results on the synthetic data sets.
%We initially discuss our experiments on the synthetic data set followed by the analysis on the real world data set.

\subsection{Data sets with synthetic annotations}

We use the two wine quality data sets (red and white wine data sets) \cite{cortez2009modeling} available in the UCI data repository \cite{asuncion2007uci}.
Each data set provides 11 attributes for each entry and a quality score between 0-10 (10 being the best). In pairwise comparison between two entries $O_i$ and $O_j$, we say that the ground truth is $z_{ij} = 1$ if $O_i$ has a higher quality score than $O_j$. 
Below we provide a short description of synthetic creation of noisy annotator labels from this data set followed by a set of three experiments investigating the reliability inference for each annotator and the effect of quality and number of annotators.\\

\noindent{\bf Creating synthetic noisy annotations}:
Given the number of annotators $K$, we create synthetic noisy annotations for the $k^{\text{th}}$ annotator by flipping the ground truth $z_{ij}^*$ based on a Bernoulli variable.
The parameter of the Bernoulli variable for annotator $k$ is denoted by $b_k$ and a higher $b_k$ implies higher chances of $z_{ij}^*$ being flipped. 
In the first experiment presented in the next section, we investigate the relation between $b_k$ used for each annotator and the probability of flipping $r_k$ determined by our joint annotator models. \\

\vspace{-2mm}
\subsubsection{Relationship between probability of flipping and annotator noise}
In this experiment, we use a set of 6 noisy annotators with $b_k = k/20$.
That is the first annotator is the best annotator with only 5\% chance of flipping where as the sixth annotator has a 30\% chance of flipping.
We train the Joint Annotator Model (JAM) and Variable Reliability Joint Annotator Model (VRJAM).
Table \ref{prob_flip} shows the values for $r_k$ estimated using JAM and the mean value of vector $\bm R_k$ estimated using VRJAM on the red wine data set (similar patterns are observed for white wine data set). Higher values for $r_k$ and mean of $\bm R_k$ imply that the annotator $k$ is inferred to be more noisy.
We also show the model accuracy in inferring the ground truth $z_{ij}^*$ over all pairs of objects in the data set in Table \ref{comparison_results}.

\begin{table}[t]
\centering
\caption {\hspace{-2mm} Values of $r_k$ \& mean($\bm R_k$) obtained on the red wine data set.}
\begin{tabular}{ll|cc}
\hline
Model & Parameter &  Values for $k=1..6; b_k=k/20$\\ \hline
JAM&$r_k$ &\ \{.032, .086, .176, .196, .246, .273\}\\
VRJAM& Mean($\bm R_k$)&  \{.033, .085, .175, .196, .245, .273\} \\\hline
\end{tabular}
\label{prob_flip}
\end{table}

\begin{table}[t]
\centering
\caption {Accuracy in inferring $z_{ij}^*$ in the synthetic data sets.}
\begin{tabular}{l|cccc}
\hline
Data set & MV & IAM & JAM & VRJAM \\ \hline
Red wine &95.9&55.2&97.9&98.0 \\
White wine &96.1&55.3&97.9&98.1  \\ \hline
\end{tabular}
\label{comparison_results}
\end{table}

From the Table \ref{prob_flip}, we observe that as the noise increases over annotators, our model successfully infers a higher probability of flipping. The values $r_k$ and the mean of vector $\bm R_k$ are fairly close to each other indicating that the JAM and VRJAM model are very similar in inferring probability of flipping. 
This is expected as VRJAM differs from JAM only in determining cluster-wise probabilities and their average should be fairly close to $r_k$. 
From Table \ref{comparison_results}, we observe that the proposed models outperform Majority Vote (MV) and Independent Annotator Modeling (IAM).
The difference in performance of JAM and VRJAM is not significant. 
This stems from the choice of synthetic annotation generation as the noise added to the annotations is uniform and does not change based on the pair of objects at hand.
Therefore VRJAM has no particular modeling advantage over the JAM scheme. 
Also, the performance of IAM is particularly low.
Our investigation reveals that the performances of the individual annotator SVR models ($\mathit f_k$ in section \ref{sec:IAM}) were very low (e.g., varied between 53.0\%-64.4\% in red wine data set) . 
Since IAM performs a sum of $f_k$ over these fairly weak models, the final performance is poor.
This shows that the IAM performance is contingent upon the model choice and can improve with a better choice for $f_k$.
However, an interesting point to note here is that the IAM performance (e.g., 55.2\% for red wine data set) lies between the performance of the best annotator ($64.4\%$ for red wine data set) and the worst annotator ($53.0\%$ for red wine data set).
This reflects the fact that IAM is susceptible to performing below collective knowledge of the crowd and can perform worse than the best available annotator. \\
 
\vspace{-2mm}
\subsubsection{Relationship between model performances and annotator noise}
\begin{figure}[t]
\begin{center}
\begin{tabular}
{@{\hspace{-0.0cm}}c@{\hspace{0.1cm}}c@{\hspace{0.0cm}}}
\raisebox{-.5\totalheight}{\begin{overpic}[trim=0.2cm .8cm 0.7cm 0.4cm,clip=true,scale=0.49]{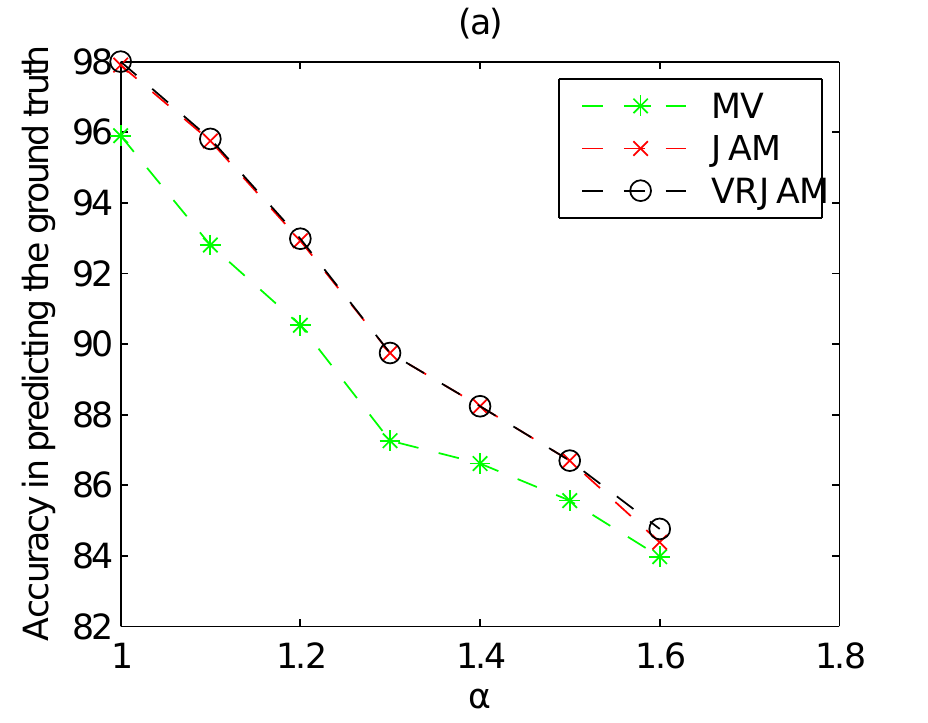}
\put (16,5) {\scriptsize{Red wine data set}}
\end{overpic}} &
\raisebox{-.495\totalheight}{\begin{overpic}[trim=0.2cm 0.8cm 0.7cm 0.4cm,clip=true,scale=0.49]{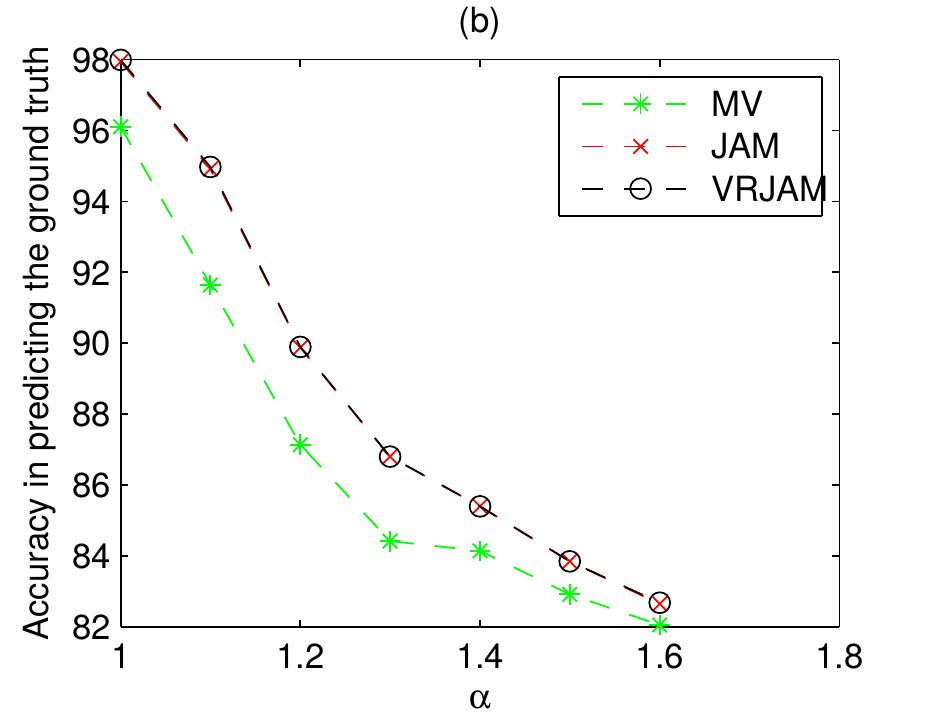}
\put (16,5) {\scriptsize{White wine data set}}
\end{overpic}
}\\
\raisebox{-.5\totalheight}{\includegraphics[trim=0.2cm 0cm 0.7cm 0.1cm,clip=true,scale=0.485]{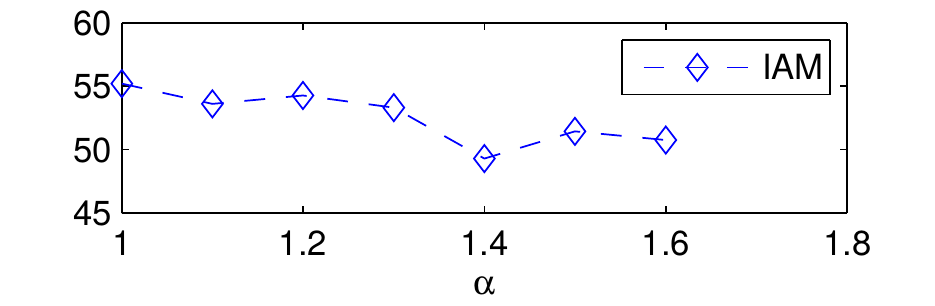}} &
\raisebox{-.495\totalheight}{\hspace{-1mm}\includegraphics[trim=0.2cm 0cm 0.7cm 0.1cm,clip=true,scale=0.485]{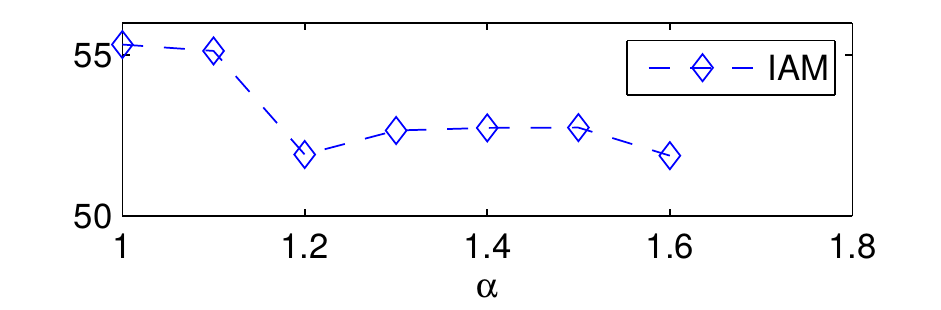}}
\end{tabular}
\end{center}
\vspace{-3mm}
\caption{Model performances with increasing annotator noises.}
\label{fig:noise_perf}
\end{figure}

In this section, we perform multiple experiments similar to the one mentioned in the previous section.
We chose a set of 6 annotators and in each experiment, we increase the parameter $b_k$.
Within an experiment, $b_k$ for the annotator $k$ is set at $\alpha k/20$ and the parameter $\alpha$ is increased by 10\% over consecutive experiments.
We plot the accuracy of the MV, IAM, JAM and VRJAM algorithms in inferring the ground truth $z_{ij}^*$ with increasing $\alpha$ in Figure \ref{fig:noise_perf}.
From the figure, we note that the model performance drops as the annotator noise increases.
Performances of the VRJAM and JAM schemes are again similar because of the reasons stated in the previous section.
Another interesting observation is that the performances of MV, JAM and VRJAM converge as the annotator noise increases.
This indicates that the joint models are likely to perform better than MV with better quality annotators.
The IAM performances are again low attributed to weak annotator modeling by the SVRs.\\

\begin{figure}[t]
\begin{center}
\begin{tabular}
{@{\hspace{-0.0cm}}c@{\hspace{0.1cm}}c@{\hspace{0.0cm}}}
\raisebox{-.5\totalheight}{\begin{overpic}[trim=0.2cm 0.8cm 0.7cm 0.4cm,clip=true,scale=0.49]{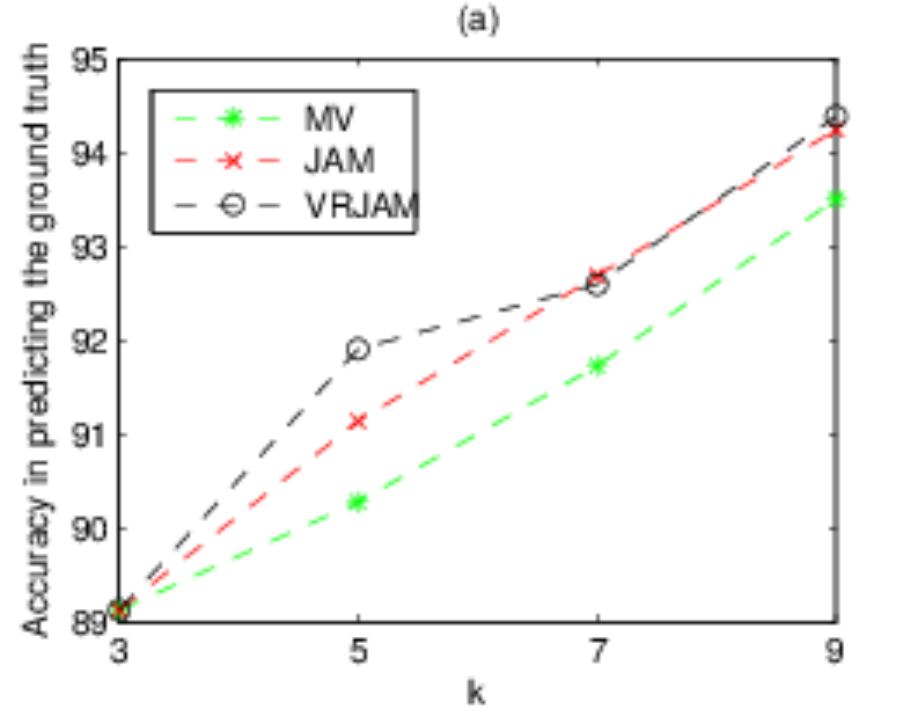}
\put (36,5) {\scriptsize{Red wine data set}}
\end{overpic}} &
\raisebox{-.495\totalheight}{\begin{overpic}[trim=0.2cm 0.8cm 0.7cm 0.4cm,clip=true,scale=0.49]{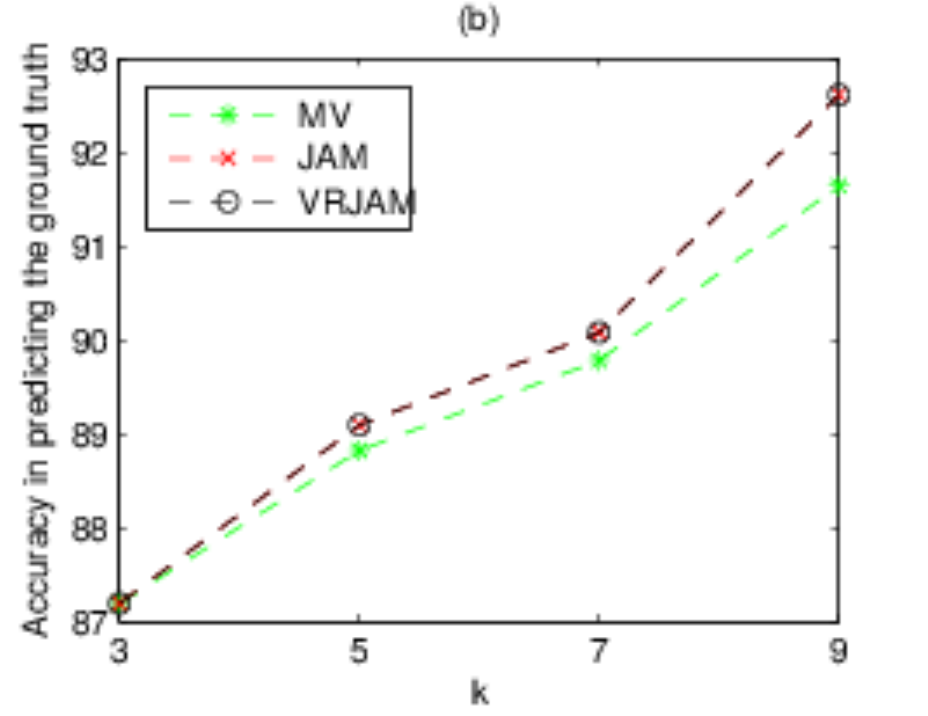}
\put (36,5) {\scriptsize{White wine data set}}
\end{overpic}}\\
\raisebox{-.5\totalheight}{\includegraphics[trim=0.2cm 0cm 0.7cm 0.1cm,clip=true,scale=0.485]{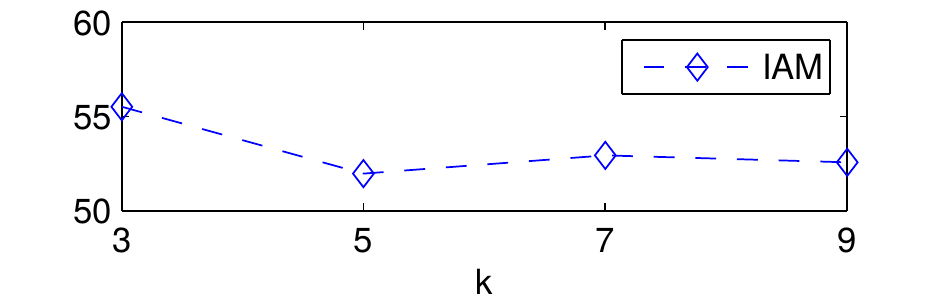}} &
\raisebox{-.495\totalheight}{\includegraphics[trim=0.2cm 0cm 0.7cm 0.1cm,clip=true,scale=0.485]{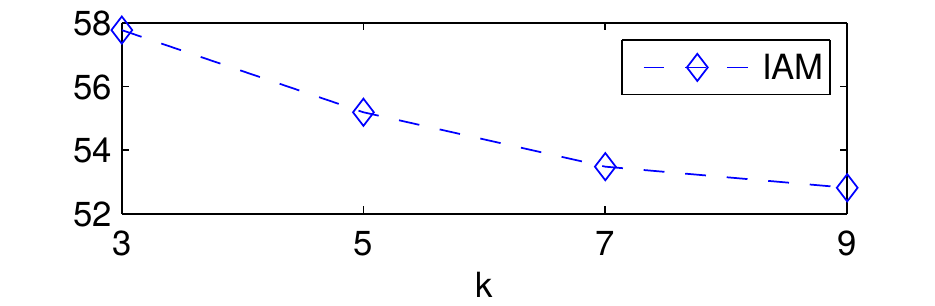}}
\end{tabular}
\end{center}
\vspace{-3mm}
\caption{Model performances with increasing number of annotators.}
\label{fig:annt_perf}
\end{figure}

\vspace{-2mm}
\subsubsection{Relationship between model performances and number of annotators}
In this section, we perform multiple experiments by varying the total count of annotators $K$. The parameter $b_k$ for the annotator $k$ is kept constant at $k/20$.
Figure \ref{fig:annt_perf} shows the plots for model performance as $K$ is varied from 3 to 9.
In this case, we observe that except for IAM, performance of all models increase with increase in number of annotators.
This indicates that addition of more noisy annotators (as $b_k<b_{k+1}$) tends to decrease IAM performance.
Also, the JAM and VRJAM models provide greater improvement over MV with addition of more annotators.
The performance of MV, JAM and VRJAM models are same at $K=3$ and the absolute improvement of the joint models over MV increases as we add more annotators.
VRJAM and JAM again perform at similar levels.
As stated, we attribute this to the nature of our synthetic labels creation where noisy annotators flip $z_{ij}^*$ solely based on $b_k$ and not based on the object attributes $\bm x_i,\bm x_j$.
%As VRJAM targets problems where annotators have variable reliability over the space spanned by $\bm x_i,\bm x_j$, the improvements over JAM are not significant.
In the next section, we test our algorithms on a data sets with machine/human annotations and analyze the results.

\subsection{Data set with machine/human annotations}
We show the results for two real world data sets, one annotated by machine experts and other by naive mechanical turk workers.
We discuss the results for these two datasets below. 

\subsubsection{Digit ranking dataset: Machine annotation}
We use a subset of the pen based recognition of handwritten digits dataset \cite{alimoglu1996combining} to rank images based on the digit value contained (for instance image with digit 9 is ranked higher than image containing any other digit).
The dataset contains 1k samples of images with 16 features, leading to ~370k possible comparisons (we do not consider comparison between images containing same values).
We initially annotate the dataset using a set of five classifier as machine annotators: K-Nearest Neighbors (KNN), Logistic Regression, Naive Bayes, Random Forest and Perceptron \cite{bishop2006pattern}.
These annotations are obtained using a 10 fold cross-validation framework.
Each classifier is trained on a subset of 3-4 features (out of 16) on 90\% of the data and results are obtained on the remaining 10\%.
This process is repeated till we annotate the entire data using the classifiers.
Note that in this dataset, we have access to the ground truth which may not always be the true (this is the case with the dataset in the next section). 
Table \ref{results_mnist} shows the performance of each classifier as a machine annotator in pairwise comparison between images. 
Table \ref{results_mnist2} shows the performance of the fusion schemes operating over the machine annotations thus obtained.
We use the entire set of 16 features in the JAM and VRJAM fusion schemes.

\begin{table}[t]
\centering
\caption{Ratio of pairwise comparisons in which a classifier ranks the image containing greater value higher than the other image in the pair. (KNN: KNN classifier, LR: Logistic Regression, NB: Naive Bayes classifier, RF: Random Forests classifier and Perc.: Perceptron).} 
\begin{tabular}{@{}l|lllll@{}}
\hline
Classifier& KNN & LR & NB & RF & Perc.\\ \hline
%& & regression & Bayes & forest& tron \\ \hline
Performance & 67.8 & 69.1 & 69.0 & 72.0 & 59.9 \\ \hline 
\end{tabular}
\label{results_mnist}
\end{table}

\begin{table}[t]
\centering
\caption {Performance of the fusion schemes on pairwise comparisons $z_{ij}^k$, as obtained from the machine annotators.}
\begin{tabular}{l|llll}
\hline 
Fusion scheme& MV & IAM & JAM & VRJAM \\
Performance & 78.0 & 65.9 & 78.1 & 79.7 \\ 
\hline
\end{tabular}
\label{results_mnist2}
\end{table}

From the Table \ref{results_mnist}, we see that the machine annotators perform in the range of 59\% to 72\% on the metric of pairwise comparison accuracy.
Results in Table \ref{results_mnist2} indicate that the MV, JAM and VRJAM schemes outperform the best machine annotator, i.e., random forests.
Where as the performances of MV and JAM are not significantly different, VRJAM performs significantly better than both MV and JAM schemes (McNemar's test \cite{trajman2008mcnemar}, significance level: 5\%, computed over the ~370k comparison samples).
This indicates that assigning a flipping probability conditioned on the pair of images at hand is essential in this data set. 
The IAM scheme again fails to beat the best annotator and performs at a value within the range of best and the worst annotator.
This indicates that an unweighted fusion of experts may perform below the collective knowledge of the crowd and weighting annotators based on individual performances may help.
In the next section, we test the fusion scheme on another real data set with human annotators.

\begin{table}[t]
\centering
\caption {Comparison of expressiveness/naturalness between TD and HFA kids. TD kids are expected to be more expressive/natural.}
\begin{tabular}{@{}l|cccc@{}}
\hline
Attribute & \multicolumn{4}{|c}{Ratio of times TD kids are inferred to } \\
 & \multicolumn{4}{|c}{have a higher rank over HFA kids} \\ \cline{2-5}
&MV&IAM&JAM&VRJAM \\\hline
Expressiveness& 64.3& 61.5&64.3&65.4\\
Naturalness& 55.7& 52.7&55.9&57.7\\
\hline
\end{tabular}
\label{safari_bob}
\end{table}

\subsubsection{Safari Bob dataset}
In this section, we test our algorithms on the Safari Bob data set \cite{dannyIS2015}.
This data set involves two populations of High Functioning Autism (HFA) and Typically Developing (TD) individuals retelling a story based on a video stimulus.
The recording of story retelling are later rated by naive Mechanical Turk (MTurk) raters for expressiveness and naturalness on a scale from 0-4 (4 being the best).
We use a set of 40 TD kids and 65 HFA kids rated by 5 annotators and infer the ground truth expressiveness and naturalness from the available ratings.
The attributes $\bm x_i$ we use to train the models are statistical functionals extracted on prosodic and spectral features from the kid's speech (mean and variance of pitch, intensity, Mel filter banks and Cepstral Coefficients) as are also used in \cite{gupta2012classification,dannyIS2015}.
Since we do not have the ground truth available for evaluation, we analyze the association of inferred expressiveness and naturalness with the population attributes of HFA and TD.
Although the relationship between autism and expressiveness/naturalness is fairly complex and undergoing extensive investigation \cite{grossman2013emotional}, TD kids are expected to be ranked higher in expressiveness/naturalness over HFA kids \cite{dannyIS2015}.
We infer the latent ground truth for expressiveness/naturalness using our models set and show (Table \ref{safari_bob}) the proportion of times the models infer TD kids to have a higher expressiveness/naturalness than HFA kids.

From the results, we observe that a TD kid is more often inferred to have a higher expressiveness/naturalness over a HFA kid.
Whereas outputs for MV and JAM are fairly close to each other, the outputs from the VRJAM has the highest proportion of times that a TD kid is inferred to be more expressive/natural than an ASD kid.
This trend is encouraging although the relation between speech expressiveness/naturalness and autism may not be this straightforward.
Due to unavailability of ground truth, this experiment can not be used to support the efficacy of proposed algorithms. However the observed results motivate the application of proposed algorithms to data sets where the ground truth is unobserved.

Overall, the experiment on synthetic, machine and human annotations in this section provide an understanding of the proposed algorithms within the aspects of annotator reliability, quality, and number of annotators. 
Although the performance of VRJAM is not significantly better in the case of synthetic annotations, results on the machine and human annotations indicate the importance of accounting for differences in the reliability of annotators based on the pair of objects at hand. 
We conclude our work in the next section and present a few future directions.

\section{Conclusion}
In this paper, we address the problem of inferring the hidden ground truth preference given noisy annotations from multiple annotators. 
We propose an EM algorithm based Joint Annotator Modeling (JAM) scheme, considering the latent ground truth preference to be a hidden variable and inferring it based on available annotation and object attributes.
Given a pair of objects, the JAM scheme infers the latent true preference order based on a set of object attributes as well as noisy annotator preferences.
The model assumes that annotators flip the true preference order based on a Bernoulli random variable and estimates annotator specific ``probability of flipping".
We further extend the model to estimate a non-constant ``probability of flipping" conditioned on the pair of objects at hand in the Variable Reliability Joint Annotator Model (VRJAM).
We test the JAM and VRJAM schemes against majority voting and Independent Annotator Modeling schemes on data sets with annotations obtained synthetically, from machines as well as from human annotators. 
Using the data set with synthetic annotations, we test the impact of annotator quality and quantity on our models. 
The results on data sets with machine annotations depicts the importance of having a variable reliability per annotator based on pair of objects at hand.
Finally, in the Safari Bob data with human annotators, we interpret the results based on the expected trends of expressiveness/naturalness in TD and HFA kids.

In the future, we aim to extend the presented algorithms by integrating other existing work in the ranking domain (e.g., active learning). 
Other work in rank aggregation inferring a rank order probability distribution can also be integrated into the proposed EM framework.
Also, within classification there are further extensions of multiple annotator models which can be incorporated into the current EM framework.
We also aim to implement the designed algorithms to other data sets such as the Safari Bob data set in understanding the diversity in perception of various psychological constructs (e.g. naturalness) by the human annotators and their relation to a target variable (e.g. autism severity).

\bibliographystyle{IEEEbib}
\bibliography{strings,refs}

\newpage
\clearpage
\appendix
{\bf Appendix 1: Proof for equation (\ref{e_step_all})}

\noindent{\it To prove}:
\begin{equation*}
\begin{aligned}
&q(z_{ij}^*)=p(z_{ij}^*|z_{ij}^1, ..,z_{ij}^K , \bm x_i, \bm x_j, \bm w, r_1, .., r_K) =\\ 
&\Big(p(z_{ij}^*|\bm x_i, \bm x_j, \bm w) \times \prod_{k=1}^K p(z_{ij}^k | z_{ij}^*, r_k)\Big) / p(z_{ij}^1, ..,z_{ij}^K) 
\end{aligned}
\end{equation*}

\noindent{\it Proof}:

\begin{equation} \label{to_prove_1}
\begin{aligned}
&q(z_{ij}^*)=p(z_{ij}^*|z_{ij}^1, ..,z_{ij}^K , \bm x_i, \bm x_j, \bm w, r_1, .., r_K) \\ 
&= p(z_{ij}^*,z_{ij}^1, ..,z_{ij}^K | \bm x_i, \bm x_j, \bm w, r_1, .., r_K) / p(z_{ij}^1, ..,z_{ij}^K)
\end{aligned}
\end{equation}

By Bayes theorem:
\begin{equation}\label{append1_bt}
\begin{aligned}
&p(z_{ij}^*,z_{ij}^1, ..,z_{ij}^K | \bm x_i, \bm x_j, \bm w, r_1, .., r_K) / p(z_{ij}^1, ..,z_{ij}^K)  \\
&= p(z_{ij}^1, ..,z_{ij}^K | z_{ij}^*, \bm x_i, \bm x_j, \bm w, r_1, .., r_K) \\ 
&\times p (z_{ij}^*| \bm x_i, \bm x_j, \bm w, r_1, .., r_K) / p(z_{ij}^1, ..,z_{ij}^K) 
\end{aligned}
\end{equation}

Based on the graphical model in Figure \ref{all_models}(c), we can say that $z_{ij}^1, ..,z_{ij}^K$ are independent of the attributes $\bm x_i, \bm x_j$ and SVR vector $\bm w$, using the ``indirect evidential effect" clause in \cite{koller2009probabilistic}. 

\begin{equation}
\begin{aligned}
p(z_{ij}^1, ..,z_{ij}^K | z_{ij}^*, \bm x_i, \bm x_j, \bm w, r_1, .., r_K) = \\
p(z_{ij}^1, ..,z_{ij}^K | z_{ij}^*, r_1, .., r_K) 
\end{aligned}
\end{equation}

Next, applying the ``common clause" effect \cite{koller2009probabilistic} to the graphical model in Figure \ref{all_models}(c), we can say that $z_{ij}^1, ..,z_{ij}^K$ are mutually independent given $z_{ij}^*$. Consequentially, $z_{ij}^k$ is also independent of all $r_{k^\prime}$ for all $k^\prime \neq k$ due to the ``common clause" effect.
Therefore:
 
\begin{equation}\label{append1_1} 
\begin{aligned}
p(z_{ij}^1, ..,z_{ij}^K | z_{ij}^*, r_1, .., r_K) =  
\prod_{k=1}^K p(z_{ij}^k | z_{ij}^*, r_k)
\end{aligned}
\end{equation}

We can also say that $z_{ij}^*$ is independent of $r_1, .., r_K$ when the probability distribution is not conditioned on $z_{ij}^1, ..,z_{ij}^K$ again using the ``common clause" effect \cite{koller2009probabilistic}.

\begin{equation}\label{append1_2}
\begin{aligned}
p (z_{ij}^*| \bm x_i, \bm x_j, \bm w, r_1, .., r_K) = 
p (z_{ij}^*| \bm x_i, \bm x_j, \bm w) 
\end{aligned}
\end{equation}

Replacing (\ref{append1_1}) and (\ref{append1_2}) into (\ref{append1_bt}), we obtain 
\begin{align}
&q(z_{ij}^*)=p(z_{ij}^*|z_{ij}^1, ..,z_{ij}^K , \bm x_i, \bm x_j, \bm w, r_1, .., r_K) \\
&= p (z_{ij}^*| \bm x_i, \bm x_j, \bm w) \prod_{k=1}^K p(z_{ij}^k | z_{ij}^*, r_k) / p(z_{ij}^1, ..,z_{ij}^K) 
\end{align}

\noindent{\bf Appendix 2: Proof for equation (\ref{updated_M2})}

\noindent{\it To prove}:

\begin{equation} \label{to_prove_2}
\begin{aligned}
\log p(z_{ij}^*,z_{ij}^1, ..,z_{ij}^K| \bm x_i, \bm x_j, \bm w, r_1, .., r_K)\\
=\sum_{k=1}^K \log p(z_{ij}^k| z_{ij}^*, r_k) + \log p(z_{ij}^*| \bm x_i, \bm x_j, \bm w)
\end{aligned}
\end{equation}

\noindent{\it Proof}:

Application of (\ref{append1_bt})-(\ref{append1_2}) to the left hand side of (\ref{to_prove_2}) yields the desired result.\\ 

\noindent {\bf Appendix 3: Probability distribution for optimization in equation (\ref{em_approx_p})}

The goal in the M-step of the EM algorithm in order to obtain $\bm w$ was to perform the following optimization.

\begin{equation}\label{appnd3_cost}
\begin{aligned}
&\bm w = \arg\max_{\bm w} \mathcal M_{\bm w}\\ 
&= \arg\max_{\bm w} \sum_{z_{ij}^* \in \{0,1\}} q(z_{ij}^*) \log p(z_{ij}^* | \bm x_i, \bm x_j, \bm w)
\end{aligned}
\end{equation}

Where
\begin{align}
&p(z_{ij}^* = 1 | \bm x_i, \bm x_j, \bm w) =  
\frac{\exp{\langle \bm w, \{\bm x_i - \bm x_j\} \rangle}}{1 + \exp{\langle \bm w, \{\bm x_i - \bm x_j\}\rangle}}\\
&p(z_{ij}^* = 0|\bm x_i, \bm x_j, \bm w) = 1 - p(z_{ij}^* = 1|\bm x_i, \bm x_j, \bm w)  
\end{align}

Instead, we performed the optimization in (\ref{em_approx_p}), restated below.

\begin{equation}\label{em_approx_p2}
\begin{aligned}
\bm w = \arg\min_{\bm w} \mathcal M_{\bm w} &= \arg\min_{\bm w} \Big(q(z_{ij}^* = 1) [1-\langle \bm w, \{\bm x_i - \bm x_j\} \rangle]_+  \\
&+ q(z_{ij}^* = 0) [1-\langle \bm w, \{\bm x_j - \bm x_i\} \rangle]_+ \Big) 
\end{aligned}
\end{equation}

Above optimization can be rewritten as shown in (\ref{append3_opt}).

\begin{equation}\label{append3_opt}
\begin{aligned}
\bm w &= \arg\max_{\bm w} \Big(q(z_{ij}^* = 1) (-1\times[1-\langle \bm w, \{\bm x_i - \bm x_j\} \rangle]_+)  \\
&+ q(z_{ij}^* = 0) (-1\times[1-\langle \bm w, \{\bm x_j - \bm x_i\} \rangle]_+) \Big) 
\end{aligned}
\end{equation}

%\begin{equation} \label{append3_opt}
%\begin{aligned}
%\bm w = \arg\max_{\bm w}  \sum_{z_{ij}^* \in \{0,1\}}q(z_{ij}^*) (-1\times[1-\langle \bm w, \{\bm x_i - \bm x_j\} \rangle]_+)  \\
%\end{aligned}
%\end{equation}

\begin{figure}[tb]
\centering
\includegraphics[trim=0cm 0cm 0cm 0cm,clip=true,scale=.38]{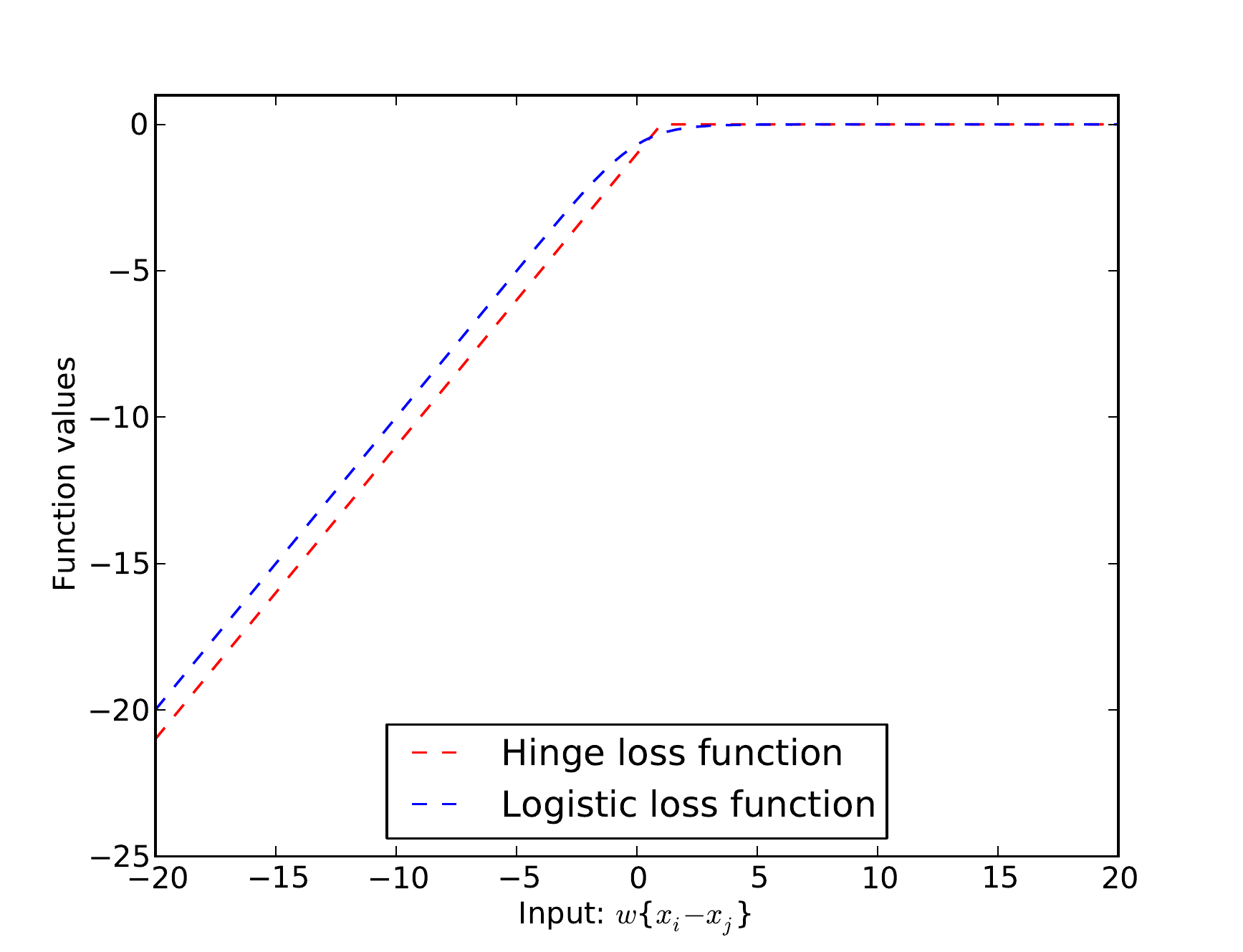}
\vspace{-4mm}
\caption{Plot comparing the values of the negative hinge loss function ($-1\times[1-\langle \bm w, \{\bm x_i - \bm x_j\} \rangle]_+$) and the log of logistic loss function ($\log p(z_{ij}^* | \bm x_i, \bm x_j, \bm w)$).}
\label{appnd3_fig}
\vspace{-5mm}
\end{figure}

We compare the negative hinge loss function $(-1 \times [1-\langle \bm w, \{\bm x_i - \bm x_j\} \rangle]_+)$ and the log of the logistic loss function ($\log p(z_{ij}^* | \bm x_i, \bm x_j, \bm w)$) stated in (\ref{appnd3_cost}).
Figure \ref{appnd3_fig} shows the values that these function take with respect to the input $\bm w \{\bm x_i - \bm x_j\}$. 
The plots indicate that the values taken by the two functions are very close to each other.
One difference is around an input value of 0, where the hinge loss function is not differentiable but the logistic loss function is.
More importantly, the slopes of the two functions are same for a large range of input and therefore, for all practical purposes, the gradient descent algorithm should provide similar results after replacing the logistic loss function with hinge loss function in the M-step of the EM algorithm.
However, we were unable to theoretically prove that the algorithm still falls under the paradigm of generalized EM algorithm, and therefore is an approximation in the EM algorithm.
%\begin{equation}
%\begin{aligned}
%=\arg\max_{\bm w} &\Big[ \Big(q(z_{ij}^* = 1) [1-\langle \bm w, \{\bm x_i - \bm x_j\} \rangle]_+  \\
%&+ q(z_{ij}^* = 0) [1-\langle \bm w, \{\bm x_j - \bm x_i\} \rangle]_+ \Big) \Big ] 
%\end{aligned}
%\end{equation}

%Since the hinge loss function is defined as
%
%\begin{equation}
%[1-\langle \bm w, \{\bm x_i - \bm x_j\} \rangle]_+ = \max \Big(0, 1-\langle \bm w, \{\bm x_i - \bm x_j\} \rangle\Big)
%\end{equation}
%
%(\ref{append3_opt}) can be written as
%\begin{equation} \label{append3_opt2}
%\begin{aligned}
%\bm w = \arg\max_{\bm w}  \sum_{z_{ij}^*}q(z_{ij}^*) \Big (-1 \times \max (0, 1-\langle \bm w, \{\bm x_i - \bm x_j\} \rangle) \Big)  \\
%= \arg\max_{\bm w}  \sum_{z_{ij}^*}q(z_{ij}^*) \Big (\min (0, \langle \bm w, \{\bm x_i - \bm x_j\} \rangle-1) \Big)  
%\end{aligned}
%\end{equation}

\end{spacing}
\end{document}